\documentclass[lettersize,journal]{IEEEtran}
\usepackage{amsmath,amsfonts}
\usepackage{algorithmic}
\usepackage{array}
\usepackage[caption=false,font=normalsize,labelfont=sf,textfont=sf]{subfig}
\usepackage{textcomp}
\usepackage{stfloats}
\usepackage{url}
\usepackage{verbatim}
\usepackage{graphicx}
\usepackage{bbm}
\usepackage{dsfont}

\usepackage{graphicx}
\usepackage{wrapfig}
\usepackage[utf8]{inputenc}
\usepackage{multirow}
\usepackage[table]{xcolor}
\usepackage{amsmath}
\usepackage{amssymb}
\definecolor{tablegray}{gray}{0.92}
\definecolor{mygreen}{rgb}{0.0, 0.5, 0.0}
\definecolor{myred}{rgb}{200,0,0}

\newcommand{\drop}[1]{\textcolor{mygreen}{($\downarrow$ #1)}}
\newcommand{\greentext}[1]{\textcolor{mygreen}{#1}}

\usepackage{booktabs}

\def\BibTeX{{\rm B\kern-.05em{\sc i\kern-.025em b}\kern-.08em
    T\kern-.1667em\lower.7ex\hbox{E}\kern-.125emX}}
\usepackage{balance}
\begin{document}
\title{RCP: Representation Consistency Pruner for Mitigating Distribution Shift in Large Vision-Language Models}
\author{Jianwei Zhang,
Chaoning Zhang,~\IEEEmembership{Senior Member,~IEEE,}
          Sihan Cao,
          Wang Liu,
          Pengcheng Zheng, Jiaxin Huang,~\IEEEmembership{Graduate Student Member,~IEEE,}
          Caiyan Qin,
          Yalan Ye,
          Wei Dong and
          Yang Yang,~\IEEEmembership{Senior Member,~IEEE}
\thanks{This work was partially supported by the National Natural Science Foundation of China under grant 62572104, and
62220106008.}
\thanks{Jianwei Zhang, Chaoning Zhang, Sihan Cao, Wang Liu, Pengcheng Zheng, Yalan Ye and Yang Yang are with the School of Computer Science and Engineering, University of
  Electronic Science and Technology of China, Qingshuihe Campus, No.~2006 Xiyuan Avenue, High-Tech Zone (West), Chengdu 611731, China (e-mail: zjw5428c@gmail.com; chaoningzhang1990@gmail.com; 2023080903002@std.uestc.edu.cn; 202552080536@std.uestc.edu.cn; zpc777@std.uestc.edu.cn; yalanye@uestc.edu.cn; yang.yang@uestc.edu.cn).}%
  \thanks{Jiaxin Huang is with the Machine Learning Department, Mohamed bin Zayed University of Artificial Intelligence, Building 1B, Masdar
  City, Abu Dhabi, United Arab Emirates (e-mail: jiaxin.huang@mbzuai.ac.ae).}%
  \thanks{Caiyan Qin is with the School of Robotics and Advanced Manufacture, Harbin Institute of Technology, Shenzhen, University Town of
  Shenzhen, Nanshan District, Shenzhen 518055, China (e-mail: qincaiyan@hit.edu.cn).}%
  \thanks{Wei Dong is with the School of Information and Control Engineering, Xi'an University of Architecture and Technology, No.~13 Yanta
  Road, Xi'an 710055, China (e-mail: dongwei156@xauat.edu.cn).}%
}

\markboth{SUBMITTED TO IEEE TRANSACTIONS ON MULTIMEDIA}%
  {Zhang \MakeLowercase{\textit{et al.}}}

\maketitle

\begin{abstract}
Large Vision-Language Models (LVLMs) suffer from prohibitive inference costs due to the massive number of visual tokens processed by the language decoder. Existing pruning methods often lead to significant performance degradation because the irreversible removal of visual tokens causes a distribution shift in the hidden states that deviates from the pre-trained full-token regime. To address this, we propose Representation Consistency Pruner, which we refer to as RCP, as a novel framework that integrates cumulative visual token pruning with a delayed repair mechanism. Specifically, we introduce a cross-attention pruner that leverages the intrinsic attention of the LLM as a baseline to predict cumulative masks, ensuring consistent and monotonic token reduction across layers. To compensate for the resulting information loss, we design a delayed repair adapter denoted as DRA, which caches the essence of pruned tokens and applies FiLM-based modulation specifically to the answer generation tokens. We employ a repair loss to match the first and second-order statistics of the pruned representations with a full-token teacher. RCP is highly efficient because it trains only lightweight plug-in modules while allowing for physical token discarding at inference. Extensive experiments on LVLM benchmarks demonstrate that RCP removes up to 88.9\% of visual tokens and reduces FLOPs by up to 85.7\% with only a marginal average accuracy drop, and outperforms prior methods that avoid fine-tuning the original model on several widely used benchmarks.
\end{abstract}


\section{Introduction}
\label{sec:introduction}

\IEEEPARstart{I}{n} recent years, Large Vision-Language Models (LVLMs)~\cite{bai2023qwen, zheng2026llava, liu2023visual, liu2024improved,
  zhu2023minigpt, du2022glm, li2023blip, zhang2023internlm, bavishi2023fuyu, team2024gemini} have achieved remarkable success across various
  multimodal tasks by combining a visual encoder~\cite{shen2025contextualcoder,radford2021learning} with a powerful large language model  ~\cite{zheng2025joint, zhang2026learning,
  touvron2023llama2, radford2021learning, achiam2023gpt, chiang2023vicuna, zheng2025lightweight}.
  This design enables unified processing of
  visual content and language instructions, and it has achieved strong performance on tasks such as visual question
  answering~\cite{zhu2016visual7w,yang2025magic}, vision-grounded dialogue~\cite{kottur2019clevr}, document
  understanding~\cite{suri2025visdom}, and multimodal reasoning~\cite{zhou2025hierarchical,lim2025vlmt,li2025vision,cao2025efficient,ding2025ra,zhang2024unleash, zheng2026towards}. As these models move from research prototypes toward broader real world deployment, improving system level efficiency and scalability
  becomes increasingly important for practical multimedia applications in real deployment settings and latency sensitive scenarios.

A typical LVLM encodes an image into a large set of visual tokens and feeds them into the language decoder together with text tokens. The resulting sequence length directly affects attention computation and memory usage, including the key value cache. As a consequence, visual tokens often become a dominant source of inference latency and serving cost. To mitigate this bottleneck, prior works~\cite{ye2025voco, huang2024ivtp, dhouib2025pact, alvar2025divprune, ye2025atp, chen2024image, zhang2024sparsevlm, zhang2025beyond, bolya2022token,cao2026languageguidedtokencompressionreinforcement,mao2025prune} have explored several representative paradigms for visual token compression and pruning. Existing methods can be typically categorized into three paradigms. Visual encoder based methods \cite{bolya2022token, chen2026evoprune} compress features inside the visual encoder to reduce redundancy early. Intermediate selection paradigms \cite{ye2025voco, alvar2025divprune, zhang2025beyond, cao2026languageguidedtokencompressionreinforcement, li2025todre} select tokens after visual encoding but before they enter the language model to construct a more compact input representation. Finally, language model internal pruning paradigms \cite{dhouib2025pact, ye2025atp, sun2025lvpruning, zhang2024jointly, zheng2023cgc} perform pruning inside the language model so that token retention can leverage richer cross modal semantic context. These paradigms differ in a fundamental trade-off. Earlier compression can save more computation but relies on weaker decision signals, while later pruning can use stronger semantic evidence after some computation has already been incurred.

 \begin{figure}[t]
 \centering
 \includegraphics[width=\columnwidth]{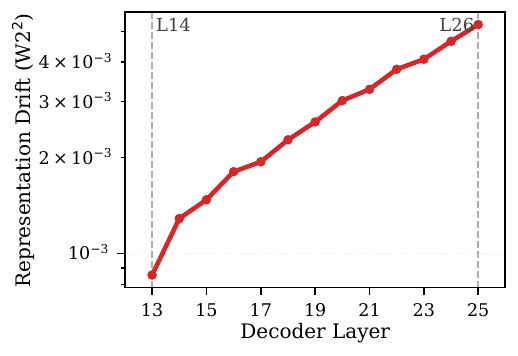}
 \caption{Layer-wise representation drift between the original and pruned models, measured by $W_2^2$.}
 \label{fig:fig_introduction_gap}
\end{figure}

Despite steady progress in reducing the number of visual tokens, existing methods often suffer noticeable performance degradation at high
  compression ratios. We argue that the key challenge is not only that fewer tokens are kept. A more fundamental issue is that pruning changes
  the representation regime encountered during decoding. Specifically, pruning can induce a mismatch between intermediate multimodal hidden
  representations produced under a pruned setting and those produced under the full-token setting, as reflected by their layer-wise
  statistical properties. Existing methods typically remove tokens that appear unimportant at the current layer, which often correlates with
  low attention scores. This removal may have only a small local effect. However, during deep decoding, repeated transformations
  and normalization operations can accumulate these modest layer-wise discrepancies into a substantial representation gap over subsequent
  layers. Figure~\ref{fig:fig_introduction_gap} illustrates this phenomenon through layer-wise representation drift measured by $W_2$. We observe that the drift is mild
  near the pruning layer but becomes progressively larger in deeper layers as decoding proceeds. This amplification also co-occurs with degradation of generation quality
  in the final outputs. These observations suggest that effective visual token compression requires not only aggressive token reduction but
  also stable representation evolution and reliable visual grounding throughout the decoding process.

Our main contributions are summarized as follows:
\begin{itemize}
\item[$\bullet$] We identify cross layer representation mismatch induced by pruning as a key factor behind performance degradation, and we provide empirical evidence that the mismatch can accumulate with depth.
\item[$\bullet$] We present the Representation Consistency Pruner (RCP), which employs a cumulative pruning strategy and a delayed repair mechanism to explicitly reduce the distribution gap between full token and pruned token representations during the pruning process.
\item[$\bullet$] We do not require any subsequent fine-tuning of the core LVLMs, because by optimizing only lightweight plug-in modules with a distribution alignment objective, we avoid costly full-model retraining and significantly ease practical deployment at scale.
\item[$\bullet$] Extensive experiments demonstrate that RCP can aggressively prune visual tokens and significantly reduce inference FLOPs, while incurring only a marginal average accuracy drop across multiple benchmarks.

\end{itemize}

\section{Related Work}
\label{sec:related}

\subsection{Large Vision--Language Models}
Recent progress in large language models (LLMs) has accelerated the development of multimodal models. Foundational research in LLMs, along with open-source initiatives, has shown that scaling and improved model architectures lead to better generalization and instruction-following capabilities~\cite{brown2020language, touvron2023llama, touvron2023llama2, radford2021learning, achiam2023gpt, workshop2022bloom, chiang2023vicuna}. Building on these advances, a family of large vision–language models (LVLMs) has emerged. These models integrate strong visual encoders with powerful LLMs, enabling tasks such as image-grounded generation and instruction following~\cite{alayrac2022flamingo, bai2023qwen, zhu2023minigpt, dai2023instructblip, li2023blip, liu2023visual, zhang2023internlm, wang2024qwen2, sun2026grasp}. Typically, LVLMs encode images or video frames using a pretrained visual encoder. The resulting visual features are then projected into the LLM’s latent space through a projection layer to align the multimodal data~\cite{li2023blip, radford2021learning, liu2023visual}. Recent engineering efforts have extended these models to support higher resolution and multiple frames, often resulting in thousands of visual tokens per example. While this improves performance on high-resolution images and document-level understanding, it also introduces substantial computational costs~\cite{bai2023qwen, wang2024qwen2, zhang2023internlm, zhang2026ghs}. These challenges have motivated increasing interest in token pruning strategies.

\subsection{Visual Token Pruning}
  Visual token pruning aims to reduce the number of visual tokens processed by LVLMs, thereby lowering inference cost and memory usage.
  Existing methods can be broadly categorized according to the stage at which pruning is performed. Encoder-side approaches operate within the
  visual encoder, such as ToMe~\cite{bolya2022token}. Interface-side methods prune tokens after visual encoding but before they are fed into
  the language model, including DivPrune~\cite{alvar2025divprune}, VisPruner~\cite{zhang2025beyond}, and
  TPRL~\cite{cao2026languageguidedtokencompressionreinforcement}. Decoder-side methods conduct pruning inside the language model after cross-
modal interaction has begun, such as PACT~\cite{dhouib2025pact}, ATP-LLaVA~\cite{ye2025atp}, and FastV~\cite{chen2024image}. Despite these
  different insertion points, most existing approaches primarily focus on identifying locally unimportant tokens, while paying limited
  attention to the representation shift introduced by pruning during subsequent decoding.

For methods acting inside the language model, a common challenge is maintaining the integrity of the feature distribution after tokens are discarded. Our approach, RCP, addresses this by inserting lightweight modules at multiple depths within the decoder. Unlike prior works that rely on independent layer-wise decisions, we introduce cumulative token masking which ensures that the set of retained tokens is monotonically refined, providing a stable information flow for the LLM. Furthermore, instead of using immediate query-aware adapters or adversarial discriminators, we propose a delayed repair strategy. By caching pruned information and applying a repair loss based on moment matching, we compensate for the distribution shift specifically in the answer generation phase. This strategy ensures that high-level reasoning is preserved while achieving aggressive token reduction in earlier stages.

\section{Methodology}
\label{sec:methods}

\begin{figure*}[!t]
\centering
\includegraphics[width=\textwidth]{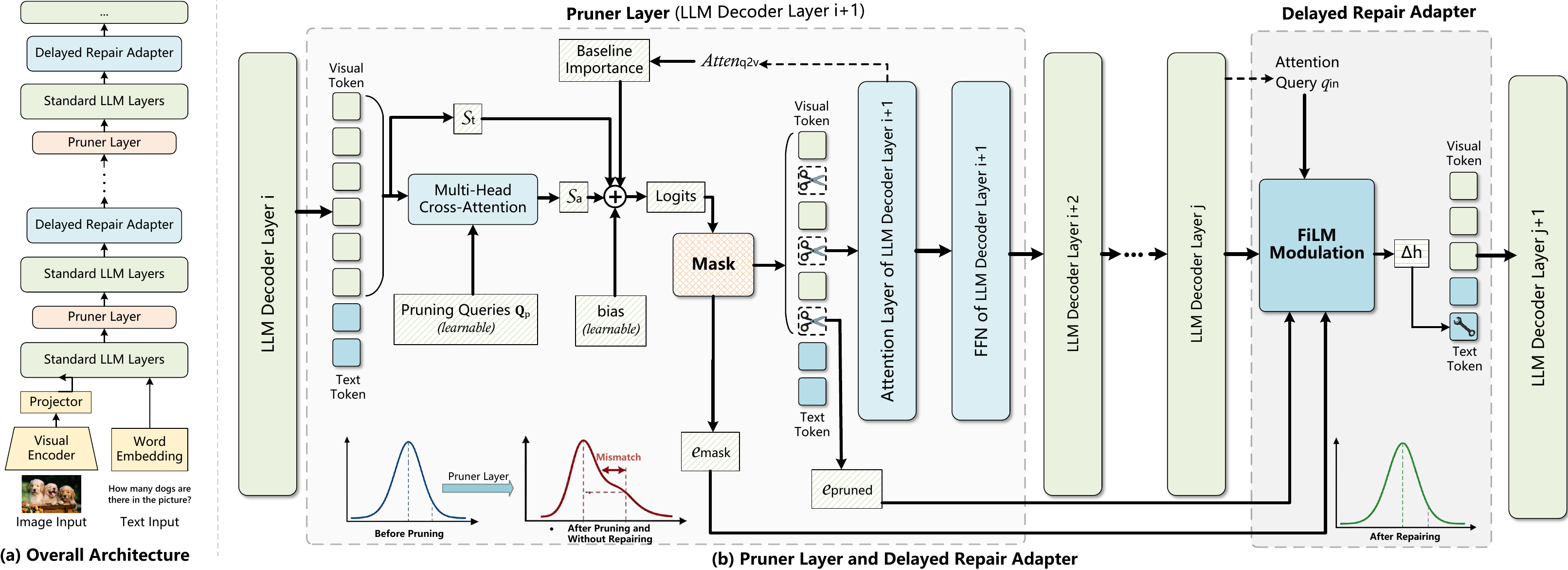}
\caption{Overview of the proposed Representation Consistency Pruner framework. (a) Overall Architecture illustrates the global system where one or more pruning modules precede a delayed repair adapter separated by multiple language model decoder layers. (b) Pruner Layer \& Delayed Repair Adapter provides the structural details for the internal components of both the pruning and repair modules.}
\label{RCP_architecture}
\end{figure*}

\subsection{Overview}
\label{sec:overview}
We propose the Representation Consistency Pruner framework, which we refer to as RCP, to reduce the computational burden of Large Vision Language Models. Since the transformer layers in the language decoder possess varying levels of semantic density, we adopt an interleaved design. We place pruning and repair modules at different depths, separating them with standard decoder blocks as illustrated in Fig.~\ref{RCP_architecture}. We provide a detailed explanation of the framework mechanics in the subsequent parts of this section. We first describe the residual cross-attention pruner in Sec.~\ref{sec:pruning}. Then, we explain the context encoding and the delayed repair adapter in Sec.~\ref{sec:repair}. Finally, we present the training objectives and the repair loss in Sec.~\ref{sec:training}. Through this integrated design, we achieve a robust balance between inference speed and reasoning quality.

\subsection{Residual Cross-Attention Pruner}
\label{sec:pruning}
As shown in Fig.~\ref{RCP_architecture}, our framework first prunes redundant visual tokens to reduce the subsequent computation. Specifically, we select a subset of the LLM decoder layers and execute the token pruning operation prior to the self-attention mechanism within these chosen layers. To achieve this, we introduce a residual cross-attention pruner aimed at identifying and discarding visual tokens that lack significant contribution to the multimodal reasoning task. The pruner utilizes the intrinsic attention weights of the frozen language model as a baseline. We determine the importance of each token at layer $\ell$ by first computing the individual scoring components. We compute the cross-attention score $S_a$ by conditioning the pruning queries on the question context.
\begin{equation}
    Q_p' = Q_p +  \frac{\sum_i h_q^{(i)}}{L_q},
\end{equation}
\begin{equation}
    S_a = \text{Aggregate}\left(\text{softmax}\left( \frac{Q_p' K_v^\top}{\sqrt{d}} \right)\right).
\end{equation}
Here $Q_p \in \mathbb{R}^{N_q \times d}$ is a matrix of $N_q$ learnable pruning queries and $K_v \in \mathbb{R}^{N \times d}$ represents the keys for the $N$ visual input tokens, while $h_q^{(i)}$ denotes the hidden state of the $i$-th question token. The value $L_q$ represents the actual length of the question sequence excluding any padding tokens. The question summary is broadcast and added to each pruning query to provide explicit question conditioning. The dimension $d$ denotes the key dimension used in the scaled dot-product attention. The function $\text{Aggregate}(\cdot)$ denotes a learnable aggregation operation over the $N_q$ pruning queries, reducing the intermediate $N_q \times N$ attention matrix to a single per-token attention score vector $S_a \in \mathbb{R}^N$.

We derive the per-token score $S_t \in \mathbb{R}^N$ by applying an MLP to each visual token independently.
\begin{equation}
    S_t = \text{MLP}(\text{proj}(h_v)),
\end{equation}
where $h_v$ represents the hidden states of the visual tokens. We derive the attention score $A \in \mathbb{R}^N$ by extracting intrinsic attention weights from the frozen language model and transforming them into the logit space. Concretely, let $a_i$ denote the aggregated attention weight for the $i$-th visual token, which is obtained from the decoder self-attention matrix by selecting the sub-block from question tokens to vision tokens and aggregating over attention heads and question-token positions. The attention score $A_i$ for the $i$-th visual token is defined as:
\begin{equation}
    A_i = \log(a_i) - \frac{\sum_{j=1}^N \tilde{m}_j \log(a_j)}{\sum_{j=1}^N \tilde{m}_j},
\end{equation}
where $N$ is the total number of visual tokens, and $\tilde{m}_j$ represents the cumulative keep mask indicating whether the $j$-th token is retained. The second term computes the masked average over the retained tokens, which avoids bias from previously discarded tokens. After determining these components, we combine the pre-trained signals and learned corrections into the final retention logit score $\text{logits} \in \mathbb{R}^N$. We add a learnable bias to adjust the logit scale:
\begin{equation}
    \text{logits} = A + S_a + S_t + \text{bias}.
\end{equation}
To enable the pruning process across layers, we define a discrete masking mechanism.
\begin{equation}
    y_{\text{soft}} = \sigma\left(\frac{\text{logits} + \epsilon}{\tau}\right),
\end{equation}
\begin{equation}
    m = \mathbb{I}[y_{\text{soft}} > 0.5] - \text{sg}(y_{\text{soft}}) + y_{\text{soft}},
\end{equation}
where $y_{\text{soft}}$ represents the differentiable mask obtained via the Gumbel-Sigmoid function, $\epsilon$ is the logistic noise, and $\tau$ denotes the temperature parameter. The function $\text{sg}(\cdot)$ denotes the stop-gradient operator. To overcome the non-differentiability of the binary mask $m$ during backpropagation, we employ the Straight-Through Estimator. During training, the exact discrete mask is applied in the forward pass, while the gradients are approximated via $y_{\text{soft}}$ in the backward pass. During inference, we discard the logistic noise and use a deterministic threshold to obtain the binary mask:
\begin{equation}
    m = \mathbb{I}\left[\sigma\left(\frac{\text{logits}}{\tau}\right) > 0.5\right].
\end{equation}
To ensure consistent pruning across layers, we treat $m$ as the layer-wise keep mask and maintain a cumulative keep mask $\tilde{m}^{(\ell)}$, updated as $\tilde{m}^{(\ell)} = \tilde{m}^{(\ell-1)} \odot m^{(\ell)}$, which prevents previously discarded tokens from reappearing in deeper layers.

\subsection{Delayed Repair Adapter}
  \label{sec:repair}
 While the residual cross attention pruner reduces computational overhead, irreversible token removal inevitably alters the hidden state distribution. To mitigate this pruning induced representation drift, we introduce the Delayed Repair Adapter. Because the representation gap amplifies in deeper layers, our method caches compact pruning context and applies it at designated repair layers. Architecturally, the adapter functions as a lightweight inter layer module inserted between consecutive decoder layers to correct hidden states.

To perform context-aware repair, the adapter must understand both the spatial distribution of the retained tokens and the semantic essence of the discarded tokens. Therefore, at each pruning layer, we extract and cache two types of compact representations: namely, a mask embedding $e_{\text{mask}} \in \mathbb{R}^d$ and a pruned feature embedding $e_{\text{pruned}} \in \mathbb{R}^d$, where $d$ denotes the hidden dimension of the underlying LLM. We construct $e_{\text{mask}}$ by using a learnable query $q \in \mathbb{R}^d$ to attend over the positional encodings $P \in \mathbb{R}^{N \times d}$ modulated by the previously obtained binary keep mask $m \in \mathbb{R}^N$. Simultaneously, we summarize the lost semantic information by computing a masked average of the hidden states $h_i \in \mathbb{R}^d$ of the discarded visual tokens using the inverted mask $(1-m)$. These cached representations are then formulated as follows:
\begin{equation}
    e_{\text{mask}} = \text{proj}\left( \text{softmax}(q \cdot (m \odot P)^\top) \cdot (m \odot P) \right),
\end{equation}
\begin{equation}
    e_{\text{pruned}} = \text{proj}\left( \frac{\sum_i (1-m_i) h_i}{\sum_i (1-m_i)} \right),
\end{equation}
where $\text{proj}(\cdot)$ denotes a linear projection layer. During the forward pass, let $x \in \mathbb{R}^{L \times d}$ denote the intermediate hidden states of sequence length $L$ outputted by the preceding LLM decoder layer. We treat the entire sequence representation $x$ directly as the query sequence $q_{in} \in \mathbb{R}^{L \times d}$ to dynamically retrieve visual context specifically for the answer tokens. We then fuse it with the globally broadcasted cached context to form a sequence-level conditioning matrix $\text{cond} \in \mathbb{R}^{L \times d}$:
\begin{equation}
    \text{cond} = e_{\text{mask}} + e_{\text{pruned}} + \text{QueryProj}(q_{in}).
\end{equation}
The repair module then uses feature-wise linear modulation to generate token-specific scale $\gamma$ and shift $\beta$ from the conditioning matrix $\text{cond}$:
\begin{equation}
    \gamma = 1 + W_\gamma \cdot \text{cond},
\end{equation}
\begin{equation}
    \beta = W_\beta \cdot \text{cond},
\end{equation}
where $W_\gamma$ and $W_\beta$ are learnable weight matrices. These modulation parameters are applied to a bottleneck transformation to obtain a residual correction $\Delta h \in \mathbb{R}^{L \times d}$:
\begin{equation}
    \Delta h = \alpha \cdot \text{Up}(\gamma \odot \text{GELU}(\text{Down}(x)) + \beta),
\end{equation}
where $\text{Down}(\cdot)$ and $\text{Up}(\cdot)$ denote the linear down-projection and up-projection layers of the bottleneck structure, respectively, which map the hidden dimension $d$ to a smaller bottleneck dimension and vice versa. We formulate the scaling factor as $1 + W_\gamma \cdot \text{cond}$ to establish an identity mapping at initialization, which preserves the pre-trained knowledge and stabilizes early training. Additionally, the learnable gating scalar $\alpha$ adaptively regulates the overall magnitude of the injected residual correction. Finally, we apply this residual correction exclusively to the answer tokens using a binary region mask $\mathbf{g} \in \{0, 1\}^L$. Specifically, the elements of $\mathbf{g}$ are set to 1 for answer-generation positions and 0 for non-answer positions, ensuring that representations are modulated only where they directly determine the final output distribution:
\begin{equation}
    h_{out} = x + \mathbf{g} \odot \Delta h.
\end{equation}
Consequently, $h_{out} \in \mathbb{R}^{L \times d}$ constitutes the final repaired hidden states, effectively bridging the pruning-induced representation gap before the sequence is further processed by the subsequent decoder layers.

\subsection{Training and Optimization}
\label{sec:training}

We design the training objective to align the representations of the compressed pruned model with those of the original full-token model
  while enforcing a target sparsity level. Since the backbone is already well optimized for multimodal tasks, we keep its parameters frozen
  and update only the pruning and repair modules. The overall training signal consists of three components: a task loss for next-token
  prediction to preserve downstream performance, a representation alignment loss to guide the delayed repair process toward the full-token
  regime, and a sparsity regularizer to control the retained token budget.

We compute the repair loss under \emph{teacher forcing} and restrict the alignment to the answer-generation region, where repair is applied. For a given decoder layer $\ell$, we collect the selected hidden states into a matrix $H \in \mathbb{R}^{T \times D}$, where $T$ is the number of tokens included in the masked region and $D$ is the hidden dimension. We then compute feature-wise statistics over token positions:
\begin{equation}
    \mu(H) = \mathbb{E}[h],
\end{equation}
\begin{equation}
    v(H) = \mathbb{E}[h \odot h] - \mu(H) \odot \mu(H),
\end{equation}
where the expectation is taken over token positions and $\odot$ denotes element-wise multiplication. Here, $\mu(H)\in\mathbb{R}^{D}$ is the per-dimension mean vector and $v(H)\in\mathbb{R}^{D}$ is the per-dimension second central moment, which corresponds to diagonal variance. Let $H_p$ denote the student hidden states produced by the pruned model with the repair adapter optionally enabled and let $H_o$ denote the teacher hidden states produced by the original full-token model with all tokens kept and repair disabled. We measure the representation drift between the student and teacher by matching their first moments and the square root of their second central moments (i.e.,
  standard deviations):
\begin{equation}
    \mathcal{L}_{\text{repair}}
    = \frac{1}{D}\left\lVert \mu(H_p) - \mu(H_o) \right\rVert_2^2
    + \frac{1}{D}\left\lVert \sqrt{v(H_p)} - \sqrt{v(H_o)} \right\rVert_2^2.
\end{equation}
The $\ell_2$ norms correspond to mean squared error averaged over feature dimensions. Importantly, rather than enforcing strict token-wise matching, this objective provides a softer, distribution-level constraint, affording the pruned model greater flexibility to optimize its representations. Mathematically, this formulation corresponds to the normalized squared 2-Wasserstein distance between diagonal-Gaussian approximations~\cite{monge1781memoire},
  which we use as the representation drift metric. By minimizing this objective, we reduce the feature-wise moment discrepancy between the
    student and teacher representations, while ignoring cross-feature covariance. In practice, we compute this representation drift on a set of designated layers and average it to obtain the final $\mathcal{L}_{\text{repair}}$.

To enforce sparsity, we define the retention ratio at decoder layer $\ell$ as
  \begin{equation}
      r_{\ell} = \frac{1}{N} \sum_{i=1}^N \tilde{m}_{i}^{(\ell)},
  \end{equation}
  where $\tilde{m}_{i}^{(\ell)}$ indicates whether token $i$ is retained at decoder layer $\ell$ after cumulative pruning. For layers without pruning, the
  cumulative mask is inherited from the nearest preceding pruning layer. We then define the global average retention rate across all decoder layers $\mathcal{L}
  _{\mathrm{dec}}$ as
  \begin{equation}
      \bar{r} = \frac{1}{|\mathcal{L}_{\mathrm{dec}}|} \sum_{\ell \in \mathcal{L}_{\mathrm{dec}}} r_{\ell}.
  \end{equation}
  We impose the sparsity loss
  \begin{equation}
      \mathcal{L}_{\text{sparse}} = \left| \bar{r} - r^* \right|,
  \end{equation}
  where $N$ is the number of initial visual tokens and $r^*$ is the target retention rate. We anneal $r^*$ during early training to ensure a smooth transition
  from dense to sparse regimes. Finally, we combine task, repair, and sparsity terms to form the total loss:
\begin{equation}
    \mathcal{L} = \lambda_{\text{task}}\mathcal{L}_{\text{task}} + \lambda_{\text{repair}}\mathcal{L}_{\text{repair}} + \lambda_{\text{sparse}}\mathcal{L}_{\text{sparse}}.
\end{equation}

\section{Experiments}
\label{sec:Experiments}

\subsection{Experimental Setup}

\begin{table*}[t]
\centering
\fontsize{7pt}{6.3pt}\selectfont 
\renewcommand{\arraystretch}{1.3}
\setlength{\tabcolsep}{4.5pt}
\caption{Performance comparison on multiple benchmarks with LLaVA-1.5-7B backbone.  We report absolute scores and the compression retention rates to Upper Bound model. \textbf{Acc.} means the average of normalized relative percentages accuracy across 7 benchmarks. }
\label{tab:main_results_7B}

\begin{tabular}{l ccccccc c}
\toprule
\textbf{Method} & \textbf{GQA} & \textbf{MME} & \textbf{POPE} & \textbf{SQA} & \textbf{VQA$^{\text{V2}}$} & \textbf{MMB}  & \textbf{VizWiz} & \textbf{Avg Acc. (\%)} \\
\midrule 

\rowcolor{tablegray} \multicolumn{9}{c}{\textit{Upper Bound (576 Tokens) (\textbf{100\%})}} \\ \midrule
\multirow{2}{*}{Vanilla} & 62 & 1763.7 & 85.9 & 69.5 & 78.5 & 64.3 & 50 & \multirow{2}{*}{100} \\
 & 100\% & 100\% & 100\% & 100\% & 100\% & 100\%  & 100\% & \\ \midrule

\rowcolor{tablegray} \multicolumn{9}{c}{\textit{Retain 192 Tokens \greentext{($\downarrow$ 66.7\%)}}} \\ \midrule
\multirow{2}{*}{ToMe~\cite{bolya2022token} } &54.39 & 1480.49 & 72.4 & 65.2 & 68.0 & 60.13 & -& \multirow{2}{*}{88.32 \drop{11.68}} \\
 & 87.72\% & 83.94\% & 84.28\% & 93.81\% & 86.62\% & 93.51\% & - &   \\ \midrule
\multirow{2}{*}{FastV~\cite{chen2024image} } & 52.7 & 1532.19 & 64.2 & 64.88 & 67.1 & 61.2 & 50.8 &  \multirow{2}{*}{88.71 \drop{11.29}} \\
 & 85.0\% & 86.87\% & 75.52\% & 91.34\% & 85.48\% & 95.18\% & 101.6\% &  \\ \midrule
\multirow{2}{*}{PDrop~\cite{sathiyanarayanan2025progressivedatadropoutembarrassingly}} & 57.16 & 1670.22 & 82.29 & \textbf{70.19} & 75.2 & 62.89 & 51.1 &  \multirow{2}{*}{97.07 \drop{2.93}} \\
 & 92.2\% & 94.7\% & 95.8\% & \textbf{101.0\%} & 95.79\% & 97.8\% & 102.2\% &  \\ \midrule
\multirow{2}{*}{HiRED~\cite{arif2025hired}} & 58.79 & 1645.3 & 81.94 & 68.4 & 75.0 & 62.41 & 50.1 &  \multirow{2}{*}{96.53 \drop{3.47}} \\
 & 94.83\% & 93.29\% & 96.39\% & 98.42\% & 95.54\% & 97.06\% & 100.2\% & \\ \midrule
  \multirow{2}{*}{VisionZip~\cite{yang2025visionzip}} & \textbf{59.4} & 1673.72 & 85.5 & 68.9 & 76.9 & \textbf{64.1} & \textbf{51.6} &\multirow{2}{*}{98.75 \drop{1.25}} \\
 & \textbf{95.8\%} & 94.9\% & 100.58\% & 99.14\% & 97.96\% & \textbf{99.69\%} & \textbf{103.2\%} &  \\ \midrule
  \multirow{2}{*}{DART~\cite{wen2025stop}} & 59.0 & 1758.02 & 81.94 & 69.8 & 76.8 & 63.21 & 51.1 &\multirow{2}{*}{98.57 \drop{1.43}} \\
 & 95.15\% & 99.68\% & 96.39\% & 100.43\% & 97.83\% & 98.3\% & 102.2\% &  \\ \midrule
 \multirow{2}{*}{\textbf{RCP (Ours)}} & 59.14 & \textbf{1787.7} & \textbf{85.56} & 68.52 & \textbf{77.54} & 63.61 & 50.39 &\multirow{2}{*}{\textbf{99.06} \drop{0.94}} \\
 & 95.38\% & \textbf{101.36\%} & \textbf{99.6\%} & 98.59\% & \textbf{98.78\%} & 98.93\% & 100.78\% &  \\ \midrule

\rowcolor{tablegray} \multicolumn{9}{c}{\textit{Retain 128 Tokens \greentext{($\downarrow$ 77.8\%)}}} \\ \midrule
\multirow{2}{*}{ToMe} & 52.48 & 1272.1 & 62.8 & 59.6 & 63.0 & 52.97 & 50.5 &\multirow{2}{*}{82.75 \drop{17.25}} \\
 & 84.65\% & 72.13\% & 73.11\% & 85.76\% & 80.25\% & 82.38\% & 101.0\% &  \\ \midrule
\multirow{2}{*}{FastV} & 49.6 & 1386.84 & 59.6 & 57.95 & 61.8 & 56.1 & \textbf{51.3} &\multirow{2}{*}{82.86 \drop{17.14}} \\
 & 80.0\% & 78.63\% & 69.46\% & 83.38\% & 78.73\% & 87.25\% & \textbf{102.6\%} &  \\ \midrule
\multirow{2}{*}{PDrop} & 56.06 & 1573.76 & 82.3 & \textbf{69.9} & 72.99 & 60.8 & 51.0 &\multirow{2}{*}{95.08 \drop{4.92}} \\
 & 90.42\% & 89.23\% & 95.8\% & \textbf{100.57\%} & 92.98\% & 94.55\% & 102.0\% &  \\ \midrule
\multirow{2}{*}{HiRED} & 57.29 & 1619.72 & 78.97 & 68.1 & 73.49 & \textbf{61.12} & \textbf{51.3} &\multirow{2}{*}{95.20 \drop{4.8}} \\
 & 92.41\% & 91.84\% & 92.9\% & 97.99\% & 93.62\% & \textbf{95.05\%} & \textbf{102.6\%} &  \\ \midrule
 \multirow{2}{*}{\textbf{RCP (Ours)}} & 56.89 & \textbf{1735.1} & \textbf{84.66} & 68.19 & \textbf{76.8} & 59.34 & 50.1 &  \multirow{2}{*}{\textbf{96.73} \drop{3.27}} \\
 & 91.76\% & \textbf{98.38\%} & \textbf{98.56\%} & 98.12\% & \textbf{97.83\%} & 92.29\% & 100.2\% &  \\ \midrule

\rowcolor{tablegray} \multicolumn{9}{c}{\textit{Retain 64 Tokens \greentext{($\downarrow$ 88.9\%)}}} \\ \midrule
\multirow{2}{*}{ToMe } & 48.68 & 1077.92 & 52.5 & 50.0 & 57.1 & 43.43 & 50.2 &  \multirow{2}{*}{73.34 \drop{26.66}} \\
 & 78.51\% & 61.12\% & 61.12\% & 71.94\% & 72.74\% & 67.54\% & 100.4\% &  \\ \midrule
\multirow{2}{*}{FastV} & 46.17 & 1189.69 & 48 & 51.1 & 55.0 & 47.7 & 50.8 &\multirow{2}{*}{73.88 \drop{26.12}} \\
 & 74.47\% & 67.45\% & 55.88\% & 73.53\% & 70.06\% & 74.19\% & 101.6\% & \\ \midrule
\multirow{2}{*}{PDrop} & 41.97 & 1034.35 & 55.9 & 68.6 & 69.29 & 33.09 & 50.7 &  \multirow{2}{*}{75.89 \drop{24.11}} \\
 & 67.69\% & 58.65\% & 65.08\% & 98.71\% & 88.27\% & 51.47\% & 101.4\% &  \\ \midrule
\multirow{2}{*}{HiRED} & 54.69 & 1514.58 & 72.84 & 68.2 & 69.79 & 59.83 & 50.2 & \multirow{2}{*}{91.46 \drop{8.54}} \\
 & 88.21\% & 85.88\% & 85.68\% & 98.13\% & 88.9\% & 93.04\% & 100.4\% &  \\ \midrule
\multirow{2}{*}{VisionZip} & 55.1 & 1594.3 & 76.2 & 71.79 & 72.4 & 60.1 & \textbf{52.9} &\multirow{2}{*}{94.81 \drop{5.19}} \\
 & 88.87\% & 90.4\% & 89.64\% & 103.29\% & 92.23\% & 93.47\% & \textbf{105.8\%} &  \\ \midrule
\multirow{2}{*}{DART} & 55.99 & \textbf{1671.82} & 73.13 & \textbf{69.8} & 72.49 & \textbf{60.23} & 51.6 &\multirow{2}{*}{ 94.40 \drop{5.6}} \\
 & 90.31\% & \textbf{94.79\%} & 86.03\% & \textbf{100.43\%} & 92.35\% & \textbf{93.66\%} & 103.2\% &  \\ \midrule
\multirow{2}{*}{\textbf{RCP (Ours)}} & \textbf{56.66} & 1671.4 & \textbf{80.49} & 68.17 & \textbf{74.53} & 59.29 & 49.94 & \multirow{2}{*}{\textbf{95.00} \drop{5.0}} \\
 & \textbf{91.39\%} & 94.77\% & \textbf{93.70\%} & 98.09\% & \textbf{94.94\%} & 92.21\% & 99.88\% & \\ 
\bottomrule
\end{tabular}
\label{main_results_llava}
\end{table*}

\subsubsection{Implementation Details}
\label{sec:implementation}

The training process for the RCP modules is conducted for one epoch
  using a subset of 10,000 samples from the VQAv2 dataset
  \cite{goyal2017making}. For benchmarks with dedicated training splits, we further perform task-specific adaptation on the corresponding training data. We utilize an initial learning rate of $5
  \times 10^{-5}$ combined with a cosine learning rate scheduler, where
  the minimum value reaches 0.1 times the initial rate. We anneal the
  Gumbel-Sigmoid temperature from 1.5 to 0.2 during training. The
  balancing of different objectives is achieved by setting the task loss
  weight to 1.5 and the sparsity loss weight to 200, while the repair
  loss weight is 40. The repair objective follows the W2-style
  distribution alignment described above. We set the number of pruning
  queries to 16. During training, we apply query-wise dropout with a
  rate of 0.2, randomly masking the contributions of 20\% of the pruning
  queries to improve generalization. The structural placement of the
  modules is determined by the specific pruning budget. For LLaVA-7B,
  when the target budget is set to 192 or 128 tokens, we insert the
  pruners at layers 5, 15, and 25. For a more aggressive budget of 64
  tokens, we shift the pruner locations to layers 2, 14, and 26. The
  delayed repair adapters are positioned at layers 23 and 30. For LLaVA-13B, which serves as an additional experimental setting, we place
  the pruners at layers 2, 16, and 28 for the 64-token budget and at
  layers 4, 18, and 33 for the 128- and 192-token budgets, while placing
  the delayed repair adapters at layers 24 and 36 in all cases. We use a training batch size of 24 in all experiments. Since RCP only optimizes the plug-in pruning and repair modules without fine-tuning
  the original model, the trainable parameter overhead remains modest. For LLaVA-7B, the pruners and delayed repair adapters contain 16.56M
  and 20.15M trainable parameters, respectively, while for LLaVA-13B, they contain 19.70M and 24.87M. Under the default setting, training on
  two RTX 4090 GPUs takes approximately 40 minutes for one epoch.

\subsubsection{Datasets and Metrics}
\label{sec:datasets}

To evaluate the effectiveness of the RCP framework, we conduct extensive and systematic experiments on seven widely used multimodal
  benchmarks. These datasets include GQA for visual reasoning \cite{hudson2019gqa}, MMBench for comprehensive evaluation
  \cite{liu2024mmbench}, MME for perception and cognition \cite{fu2023mme}, POPE for assessing object hallucination \cite{pham2024h},
  ScienceQA for multimodal science questions \cite{saikh2022scienceqa}, VQAv2 for visual question answering \cite{goyal2017making}, and VizWiz
  for visual accessibility tasks \cite{gurari2018vizwiz}. For benchmarks lacking dedicated training splits, such as POPE and MME, we directly
  evaluate the models trained on the VQAv2 dataset to assess their zero-shot generalization capabilities in a consistent setting. 

We use LLaVA-1.5-7B as the primary full-token upper bound, namely the original model without visual token pruning. Furthermore, we extend our evaluation to the LLaVA-1.5-13B model across the same set of benchmarks to verify the scalability and robustness of our method across different model scales. In our reporting, we provide both the absolute task performance on each benchmark and the relative percentage compared to the corresponding upper bound model to support a thorough analysis of the efficiency and accuracy trade-off. Additionally, because our framework employs an adaptive pruning strategy, the number of retained visual tokens is dynamically determined for each specific input. Therefore, we calculate and report the average token count during inference across all benchmarks. By adopting these comprehensive evaluation protocols across multiple architectures, we demonstrate the capability of our method to maintain high reasoning quality while significantly reducing token redundancy.

\subsection{Experimental Results}
\subsubsection{Main Results}
We evaluate the effectiveness of the RCP framework across various visual understanding benchmarks to showcase the impact of vision token pruning on reasoning capability. As illustrated in Table~\ref{tab:main_results_7B}, we compare the results of different pruning strategies on the LLaVA-1.5-7B model across three specific retention levels including 192 and 128 and 64 visual tokens. When retaining 192 visual tokens, our method incurs a performance drop of only 0.94\% which outperforms all other comparative methods. Our approach even surpasses the performance of the full token upper bound model on the MME benchmark while effectively maintaining the original performance levels on POPE and VQAv2 and VizWiz. When the token budget is further reduced to 128, the accuracy of our method drops by 3.27\%. At the most aggressive compression level of 64 tokens, the performance decrease is restricted to 5.0\% which also exceeds the accuracy of all other baseline techniques. These results highlight the exceptional performance of RCP under high compression ratios. More importantly, while several existing methods require fine-tuning of the language model to enhance accuracy, our framework achieves these results without any fine-tuning of the frozen backbone. The RCP framework also demonstrates strong scalability on the LLaVA-1.5-13B model where Table~\ref{tab:main_results_LLaVA13B} shows that the average performance drop remains minimal at 192 visual tokens. These consistent findings across different model scales confirm the task awareness and generalizability of the RCP architecture.

\begin{table}[t]
\centering
\fontsize{6pt}{5pt}\selectfont
\caption{Performance comparison on multiple benchmarks with LLaVA-1.5-13B backbone.}
\resizebox{\columnwidth}{!}{
\begin{tabular}{l | c | ccc | c}
\toprule
Method & Token & \textbf{GQA} & \textbf{MME} & \textbf{POPE} & \textbf{Avg.} \\
\midrule

\multirow{2}{*}{LLaVA-1.5-13B}
& \multirow{2}{*}{576}
& 63.3 & 1789.68 & 85.99 & - \\
& & 100\% & 100\% & 100\% & 100\% \\

\midrule

\multirow{2}{*}{FastV~\cite{chen2024image}}
& \multirow{2}{*}{192}
& 54.25 & 1515.86 & 64.66 & - \\
& & 85.7\% & 84.7\% & 75.2\% & 81.87\% \\

\midrule

\multirow{6}{*}{Ours}
& \multirow{2}{*}{192}
& \textbf{60.20} & \textbf{1786.10} & \textbf{85.73} & - \\
& & \textbf{95.11\%} & \textbf{99.8\%} & \textbf{99.7\%} & \textbf{98.20\%} \\
\cmidrule{2-6}
& \multirow{2}{*}{128}
& \textbf{58.31} & \textbf{1742.43} & \textbf{81.66} & - \\
& & \textbf{92.12\%} & \textbf{97.36\%} & \textbf{94.96\%} & \textbf{94.81\%} \\
\cmidrule{2-6}
& \multirow{2}{*}{64}
& \textbf{57.70} & \textbf{1660.29} & \textbf{80.49} & - \\
& & \textbf{91.15\%} & \textbf{92.77\%} & \textbf{93.60\%} & \textbf{92.51\%} \\

\bottomrule
\end{tabular}
}
\label{tab:main_results_LLaVA13B}
\end{table}

\begin{table*}[t]
  \centering
  \fontsize{7pt}{6pt}\selectfont
  \caption{Efficiency analysis including cache storage memory, latency, CUDA times, and FLOPs, where $\Delta$ denotes the reduction ratio.}
  \begin{tabular}{l | c | c | cc | cc | cc}
    \toprule
    Method & Avg. Token & Relative Accuracy & Storage (MB) & $\Delta$ & CUDA Times (ms) & $\Delta$ & FLOPs (T) & $\Delta$ \\
    \midrule
    Upper Bound & 576 & 100\% & 302.4 & - & 403.1 & - & 9.6 & - \\
    FastV & 192 & 88.71\% & 100.8 & 66.7\% & 230.1 & 42.9\% & 2.0 & 79.2\% \\
   
    \midrule
     & 192 & 99.06\% & 100.8 & 66.7\% & 263.7 & 34.6\% & 3.52 & 63.3\% \\
    Ours & 128 & 96.73\% & 67.2 & 77.8\% & 196.0 & 51.4\% & 2.4 & 75.0\% \\
     & 64 & 95.00\% & 33.6 & 88.9\% & 169.6 & 57.9\% & 1.37 & 85.7\% \\
    \bottomrule
  \end{tabular}
\label{tab:efficiency_analysis}
\end{table*}

\subsubsection{Efficiency Analysis}
Table~\ref{tab:efficiency_analysis} reports an efficiency analysis of RCP in terms of cache storage and total FLOPs under different average visual-token budgets. Compared to the full-token upper bound, RCP reduces FLOPs by 63.3\%, 75.0\%, and 85.7\% when retaining 192, 128, and 64 visual tokens, respectively. The corresponding cache storage decreases proportionally from 302.4\,MB to 100.8\,MB, 67.2\,MB, and 33.6\,MB. Under the same 192-token budget, FastV achieves lower FLOPs but suffers a larger accuracy drop, whereas RCP preserves accuracy by incorporating distribution alignment. Overall, these results indicate that RCP provides a favorable efficiency--accuracy trade-off, especially under aggressive token budgets in
  practical deployment.

\subsection{Ablation and Analysis}
\label{sec:ablation}
In this section, we systematically evaluate the individual contributions of our core components to validate our architectural designs. Please note that the detailed analysis on \textbf{Hyperparameter Sensitivity} is provided in the \textcolor{red}{Supplementary Material} for further reference.

\subsubsection{Component Effectiveness}
We investigate the individual contributions of each module within the RCP framework by comparing the full system against several degraded variants on the VQAv2 and POPE benchmarks as summarized in Table~\ref{tab:ablation_components}. The results indicate that the residual pruner is a critical component because replacing it with a simple Top-K selection strategy leads to a significant performance decline. This Top-K variant selects tokens solely according to the highest
  attention scores at each stage under the pruning target, which fails to fully capture the intricate inter-modal dependencies learned by our residual scoring logic.

\begin{figure}[t]
\centering
\includegraphics[width=\columnwidth]{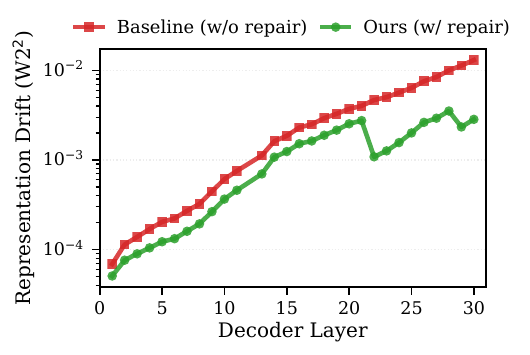}
\caption{Comparison of the repair loss between the baseline variant using only the pruner and the full RCP framework to demonstrate the stabilization effect provided by the repair adapter and the repair loss.}
\label{fig:method_comparison}
\end{figure}

To understand the stabilization effect of our training objectives, we examine the repair loss as shown in Figure~\ref{fig:method_comparison} to evaluate how closely the hidden states of the pruned model match those of the reference full token model. We compare a variant using only the pruner with task and sparsity losses against the complete RCP architecture. The full model maintains a much lower repair loss, which suggests that the combination of the delayed repair adapter and the alignment loss based on moment matching effectively mitigates the representation shift. Furthermore, the comparison in Table~\ref{tab:ablation_components} between the mean only repair loss and the full objective confirms that second order statistics are necessary for complete distribution restoration. In the mean-only variant, we retain only the mean alignment term $
  \lVert \mu_p - \mu_o \rVert_2^2$ and remove the standard-deviation
  alignment term in $\mathcal{L}_{\text{repair}}$, which results in less
  stable performance compared with the full alignment strategy. These findings demonstrate that the synergy between residual pruning and targeted distribution repair is the key to maintaining high reasoning fidelity under substantial token reduction.

\begin{table}[t]
\centering
\fontsize{6pt}{5pt}\selectfont
\caption{Ablation study of individual modules in the RCP framework.}
\label{tab:ablation_components}
\resizebox{\columnwidth}{!}{
\begin{tabular}{l | ccc}
\toprule
Method & VQAv2 & POPE & Avg. \\
\midrule
Upper Bound (Full) & 78.5 & 85.9 & 82.20 \\
RCP (Full) & \textbf{74.53} & \textbf{80.49} & \textbf{77.51} \\
\midrule
w/o Pruner (Top-K) & 67.13 & 70.32 & 68.73 \\
w/o Adapter & 74.13 & 79.62 & 76.88 \\
w/o Repair Loss & 72.98 & 79.03 & 76.00 \\
Mean-only Repair Loss & 73.06 & 79.42 & 76.24 \\
\bottomrule
\end{tabular}
}
\end{table}

\begin{table}[h]
\centering
\fontsize{5pt}{4pt}\selectfont
\caption{Effect of pruner quantity and layer placement.}
\label{tab:ablation_pruners}
\resizebox{\columnwidth}{!}{
\begin{tabular}{c | ccc}
\toprule
Pruning Layer Indexes & VQAv2 & POPE & Avg. \\
\midrule
{[2]} & 71.34 & 72.52 & 71.93 \\
{[2, 14]} & 74.11 & 80.21 & 77.16 \\
{[2, 14, 26]} & \textbf{74.53} & \textbf{80.49} & \textbf{77.51} \\
\midrule
{[3, 15, 27]} & 74.11 & 79.41 & 76.76 \\
{[4, 16, 28]} & 74.13 & 80.53 & 77.33 \\
\bottomrule
\end{tabular}
}
\end{table}

\subsubsection{Pruner Quantity and Placement}
We investigate the impact of the number and placement of pruning modules within the language model layers. Our results indicate that employing three pruning stages yields higher accuracy than using only one or two stages. This improvement is consistent with the intuition that a multi-stage approach enables a more progressive and adaptive reduction of visual tokens. Regarding placement, we observe that initiating pruning in earlier decoder layers is beneficial. Among the configurations evaluated, placing the pruners at layer indexes 2 and 14 and 26 achieves the best overall accuracy. This suggests that removing redundant visual information early can help subsequent layers focus more effectively on the most relevant multimodal evidence, thereby reducing interference from less informative visual tokens.
\begin{table}[t]
\centering
\caption{Target budgets compared with actual retention rates at various layer depths.}
\label{tab:retention_analysis}
\resizebox{\columnwidth}{!}{
\begin{tabular}{l ccc}
\toprule
Target & Layer 5 Rate & Layer 15 Rate & Layer 25 Rate \\
\midrule
192 Tokens & 60.21\% & 7.47\% & 1.56\% \\
128 Tokens & 25.84\% & 4.13\% & 1.54\% \\
\midrule\midrule
Target & Layer 2 Rate & Layer 14 Rate & Layer 26 Rate \\
\midrule
 64 Tokens & 16.72\% & 4.59\% & 1.44\% \\
\bottomrule
\end{tabular}
}
\end{table}

\subsubsection{Adapter Quantity and Placement}

We investigate the influence of the
  number and placement of delayed repair adapters on model performance.
  As shown in Table~\ref{tab:ablation_adapters}, employing two adapters improves performance
  compared to using a single adapter. Among the configurations
  evaluated, placing the adapters at post layer indexes 23 and 30
  achieves the best overall accuracy. This suggests that delaying repair
  for several layers is more effective than applying it immediately
  after pruning. These deeper layers are
  closer to answer-token prediction and thus more directly affect the
  final outputs, making repair at later
  stages more beneficial.

\begin{table}[t]
\centering
\fontsize{5pt}{4pt}\selectfont
\caption{Ablation of adapter quantity and placement.}
\label{tab:ablation_adapters}
\resizebox{\columnwidth}{!}{
\begin{tabular}{c | ccc}
\toprule
Post Layer Indexes & VQAv2 & POPE & Avg. \\
\midrule
{[23]} & 74.46 & 77.73 & 76.10 \\
\midrule
{[23, 30]} & \textbf{74.53} & 80.49 & \textbf{77.51} \\
\cmidrule{1-4}
{[22, 29]} & 74.42 & \textbf{80.52} & 77.47 \\
{[14, 26]} & 74.27 & 80.23 & 77.25 \\
\bottomrule
\end{tabular}
}
\end{table}

\subsubsection{Retention Rate Analysis}
We analyze how the retained visual-token budget evolves across layers. The results in Table~\ref{tab:retention_analysis} show that retention rates decrease monotonically as decoding proceeds to deeper layers, indicating that the pruners progressively filter redundant features at each pruning stage.

\begin{figure}[!t]
\centering
\includegraphics[width=\columnwidth]{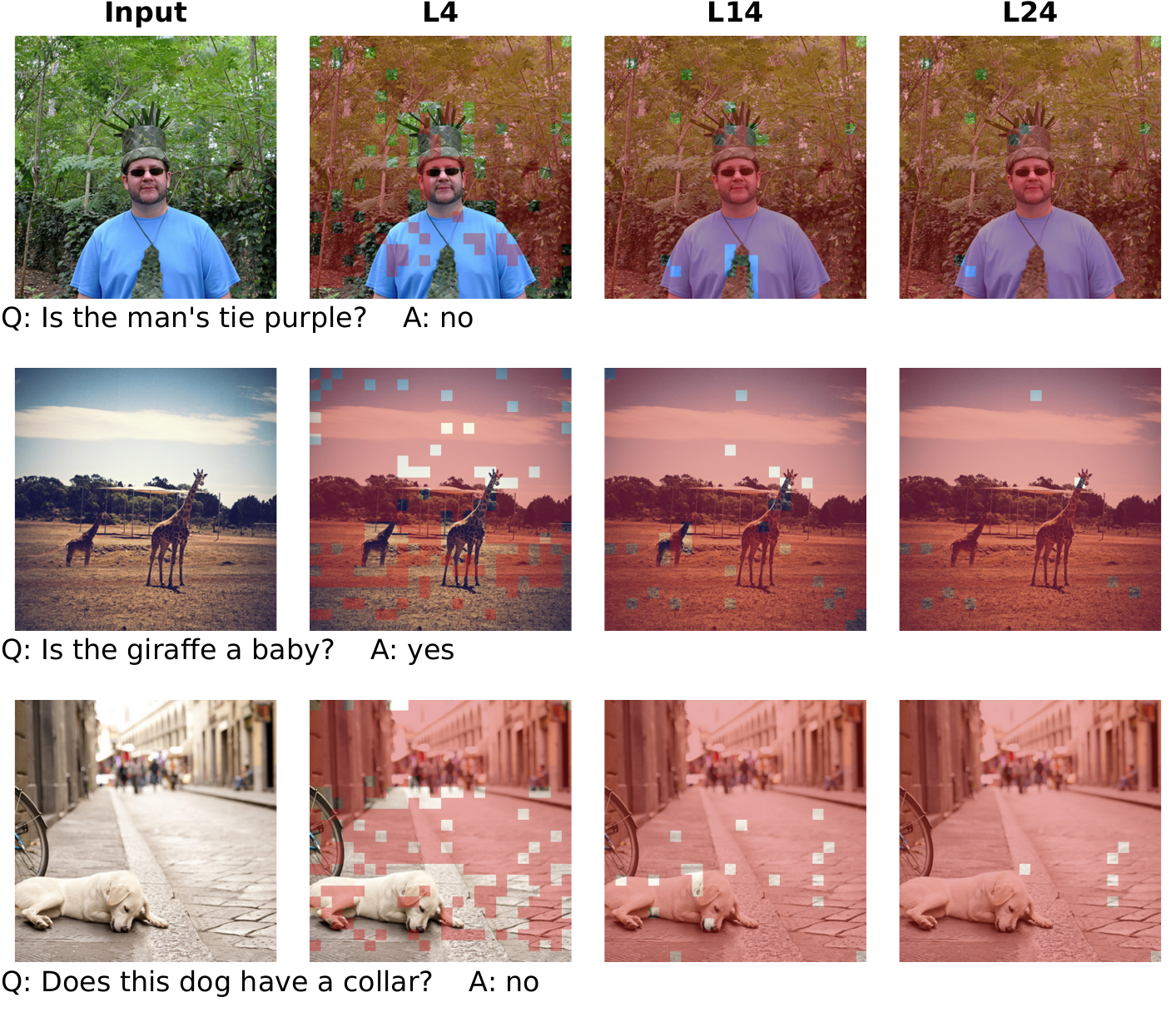}
\caption{Visualization of pruning masks at Layer 5, Layer 15, and Layer 25 under the 128 token setting which shows the progressive transition from background suppression to salient region focus.}
\label{fig:Visualization}
\end{figure}

\subsubsection{Visualization Results}

Figure~\ref{fig:Visualization} shows the pruning masks under the 64 token setting at Layer 5, Layer 15, and Layer 25. Early pruning at Layer 5 is relatively conservative and mainly removes scattered background patches. As the pruning stages go deeper, the retained tokens become more concentrated on salient regions, while
redundant areas are increasingly suppressed. The visualization suggests that multi-stage pruning yields a progressive and stable reduction of visual tokens.

\section{Conclusion}
\label{sec:conclusion}

This paper presents the Representation Consistency Pruner (RCP), a novel framework designed to resolve the distribution shift caused by visual token removal in Large Vision-Language Models (LVLMs). Our methodology introduces a cumulative residual pruning strategy combined with a delayed repair mechanism to compensate for information loss during the answer generation stage. By utilizing a repair loss based on moment matching, the framework enables the student model to synchronize its feature statistics with a full token teacher model without the need for expensive fine-tuning. Extensive evaluations across multiple benchmarks demonstrate that our method significantly improves inference efficiency, while maintaining high reasoning quality. These findings provide a practical and efficient path for the deployment of large multimodal models on resource constrained devices.

\bibliographystyle{IEEEtran}
\bibliography{main}

\end{document}